\documentclass[runningheads]{llncs}
\usepackage{graphicx}
\usepackage{tabularx} 
\usepackage{amsmath}  
\usepackage{amsfonts}
\usepackage{subcaption}
\usepackage{hyperref}
\usepackage{makecell}

\usepackage{array}
\usepackage{arydshln}
\usepackage{booktabs}

\begin{document}
\title{Enabling Synthetic Data adoption in regulated domains
}
%
%

\author{Giorgio Visani\inst{1,2} \and
Giacomo Graffi\inst{2} \and
Mattia Alfero\inst{2} \and
Enrico Bagli\inst{2} \and Davide Capuzzo\inst{2} \and
Federico Chesani\inst{1}}
\authorrunning{Visani et al.}
%
\institute{University of Bologna, School of Informatics \& Engineering, viale Risorgimento 2, 40136 Bologna, Italy, \;
\email{giorgio.visani@unibo.it} \and
CRIF S.p.A., via Mario Fantin 1-3, 40131 Bologna, Italy}

\maketitle              
\begin{abstract}
The switch from a Model-Centric to a Data-Centric mindset is putting emphasis on data and its quality rather than algorithms, bringing forward new challenges. In particular, the sensitive nature of the information in highly regulated scenarios needs to be accounted for. Specific approaches to address the privacy issue have been developed, as Privacy Enhancing Technologies. However, they frequently cause loss of information, putting forward a crucial trade-off among data quality and privacy. A clever way to bypass such a conundrum relies on Synthetic Data: data obtained from a generative process, learning the real data properties. Both Academia and Industry realized the importance of evaluating synthetic data quality: without all-round reliable metrics, the innovative data generation task has no proper objective function to maximize. Despite that, the topic remains under-explored. For this reason, we systematically catalog the important traits of synthetic data quality and privacy, and devise a specific methodology to test them. The result is DAISYnt (aDoption of Artificial Intelligence SYnthesis): a comprehensive suite of advanced tests, which sets a de facto standard for synthetic data evaluation. As a practical use-case, a variety of generative algorithms have been trained on real-world Credit Bureau Data. The best model has been assessed, using DAISYnt on the different synthetic replicas. Further potential uses, among others, entail auditing and fine-tuning of generative models or ensuring high quality of a given synthetic dataset. From a prescriptive viewpoint, eventually, DAISYnt may pave the way to synthetic data adoption in highly regulated domains, ranging from Finance to Healthcare, through Insurance and Education.
\end{abstract}

\section{Introduction}

Critical aspects of a valuable dataset are data quality and privacy. The former is stressed in the Data-Centric mindset pioneered by Andrew Ng, while the latter is required by novel regulations such as the GDPR \cite{hoofnagle_european_2019} and the U.S. FERPA \cite{gilbert_family_2007} and HIPAA \cite{cohen_hipaa_2018}, educational and medical data privacy respectively. Privacy Enhancing Technologies \cite{wang_privacy-enhancing_2009} already help protecting sensitive data, at the cost of an information loss. In fact, privacy and data quality behave as two antagonistic features. A clever way to potentially avoid such conflict relies on Synthetic Data: data obtained from a generative process, learning real data properties \cite{assefa_generating_2020}.\\
The quest for valuable synthetic data is highly relevant in regulated domains such as Finance \cite{assefa_generating_2020} and Healthcare \cite{wang_generating_2019}, where they may enable several use-cases such as: i) enforcing privacy protection, ii) facilitating data sharing among companies and towards the research community, ii) tackling class imbalance (eg. fraud detection), iv) increasing the amount of data for prediction models. Despite that, the assessment of synthetic data quality and privacy remains an under-explored, although vital, topic. Whilst few taxonomies and tests have been proposed, we feel the need for a decisive improvement.
\smallskip

In this paper we tackle the open question of how to evaluate the quality and privacy of tabular synthetic data. Firstly, we systematically catalog their most important features into three concepts: Statistical Similarity, Data Utility and Privacy. To measure these notions, we devise appropriate state-of-the-art tests yielding a numeric value in the range $[0,1]$, where higher metrics imply better performance. The final result is DAISYnt (aDoption of Artificial Intelligence SYnthesis):
a comprehensive and easy to use test suite, that sets a de facto standard for synthetic data evaluation.  As a practical use-case, a variety of generative algorithms have been trained on real-world Credit Bureau Data. The best model has been assessed, using DAISYnt on the different synthetic replicas. Further possible DAISYnt applications entail auditing and fine tuning of the models or ensuring high quality of a given synthetic dataset. 
\smallskip

In the following, Section \ref{Related Work} contains taxonomy and literature review. Section \ref{General Comparison Tests} is dedicated to general purpose tests, while Sections \ref{Distributions Comparison Tests}, \ref{Data Utility Tests} and \ref{Privacy Tests} respectively concern with distribution similarity, data utility and privacy tests. Section \ref{Application} contains DAISYnt application on Credit Scoring data, while Section \ref{Conclusions} contains a discussion on its implications and future perspectives. Methodological sections contain DAISYnt graphs and results on the Adult\footnote{The task is to predict whether a given adult makes more than 50.000\$ a year based on attributes such as education, weekly working hours, etc.} dataset from the UCI repository \cite{Dua:2019}.

\section{Related Work}\label{Related Work}

Consider to have two different datasets, $D_{train}$ and $D_{test}$ coming from the same Data Generating Process (DGP). $D_{train}$ is used to fit a generative model $G$ with the capability of producing a new dataset, $D_{synth}$. To avoid any bias, $D_{test}$ will act as the benchmark dataset.\\
Previous approaches to synthetic data evaluation are well summarized into three concepts:
\begin{itemize}
	\item $D_{synth}$ shall retain all the statistical properties of the original data, i.e. $D_{test}$.

	\item Data Utility: Prediction models built respectively on $D_{train}$ and $D_{synth}$ shall be the most similar as possible. -\textit{\small Model comparison is carried out on $D_{test}$}-

	\item Privacy of the $D_{train}$ individuals and new ones (emulated by $D_{test}$) shall be guaranteed.
\end{itemize}

\begin{figure}[htbp]
	\centering
	\includegraphics[width=0.4\textwidth]{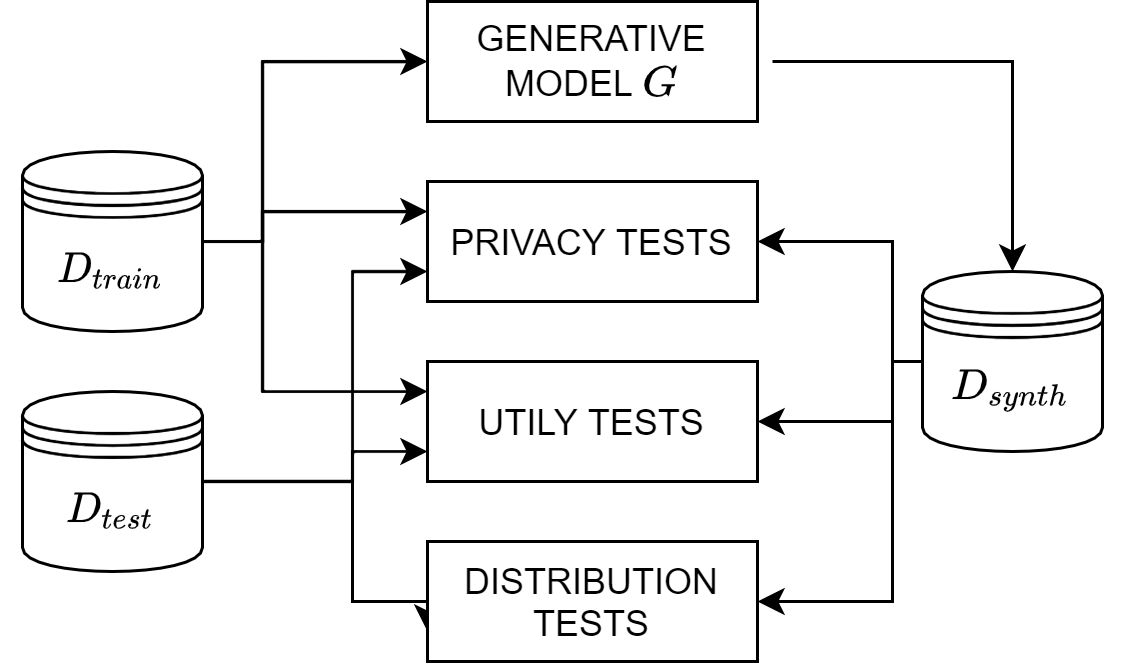}
	\caption{DAISYnt Pipeline}
	\label{data_split}
\end{figure}

Hereafter, we review statistical similarity and data utility, while the privacy topic, has its own dedicated Section \ref{Privacy Tests}. In \cite{hittmeir_utility_2019-1}, the authors compare the two correlation matrices and evaluate data utility comparing an aggregate performance metric, namely MAE. The same authors in \cite{hittmeir_utility_2019} consider dataset similarity as well, by matching the histograms of univariate distributions. Unfortunately, the comparison relied on visualization techniques, without providing any similarity metric. In \cite{wang_generating_2019}, the correlation matrix comparison is performed either, while the similarity of univariate distributions is assessed using the Kolmogorov-Smirnov (KS) two-sample test. In \cite{galloni_novel_2020} correlation is calculated by means of a particular $\phi$ coefficient (enabling its usage on both categorical and continuous variables). In \cite{jordon_measuring_2018} the Synthetic Ranking Agreement (SRA) is proposed to assess data utility, even if it is based on heuristic assumptions. In \cite{alaa_how_2021}, the authors devise three separate metrics to evaluate distribution similarity employing innovative tools. Eventually, the Synthetic Data Vault (SDV) \cite{patki_synthetic_2016} provides a sandbox for open-source generative models and related quality metrics.




\section{General Comparison Tests}\label{General Comparison Tests}

In many business scenarios, data is solely used for descriptive statistics. The following tests check $D_{synth}$ usefulness for these general-purpose, yet very popular, tasks.

\subsection{Pairwise Correlation}
Pairwise correlation is a well-known technique to measure the strength of the association between two variables. The test compares the pairwise Spearman correlation matrices $\mathbf{R}^T,\mathbf{R}^S$, respectively obtained on the $D_{test}$ and $D_{synth}$ datasets. Spearman retrieves a good amount of non-linear correlations, while it behaves as Pearson coefficient for linear dependence. However, it requires numerical variables. To this end, we encode categorical variables using the CatBoost encoding \cite{dorogush_catboost_2018}, i.e. an improved version of the Target Encoding which provides guarantees about no information leakage. 
The technique retains the relationship among the variable to be encoded and the target, while it may distort associations with other independent variables. To compensate that, the Catboost Encoder is trained on the $D_{test}$ and its mapping is applied to $D_{synth}$, ensuring the same distortion on both datasets, i.e. preserving their similarity.
\smallskip

Matrix similarity is evaluated through a rescaled version of the Frobenius norm \cite{herdin_correlation_2005}:
$$
d_{\mathrm{corr}}\left(\mathbf{R}^T, \mathbf{R}^S\right)=1-\frac{\operatorname{tr}\left\{\mathbf{R}^T \mathbf{R}^S\right\}}{\left\|\mathbf{R}^T\right\|_{F}\left\|\mathbf{R}^S\right\|_{F}} = 1-\frac{ < \mathbf{R}^T, \mathbf{R}^S>_F}{\left\|\mathbf{R}^T\right\|_{F}\left\|\mathbf{R}^S\right\|_{F}}
$$
The Frobenius inner product concatenates the rows of the matrix and computes the euclidean inner product between the two vectors. Intuitively, it consists of element-wise comparison of the two matrices.

\begin{figure}[htbp]
	\centering
	\includegraphics[width=0.7\textwidth]{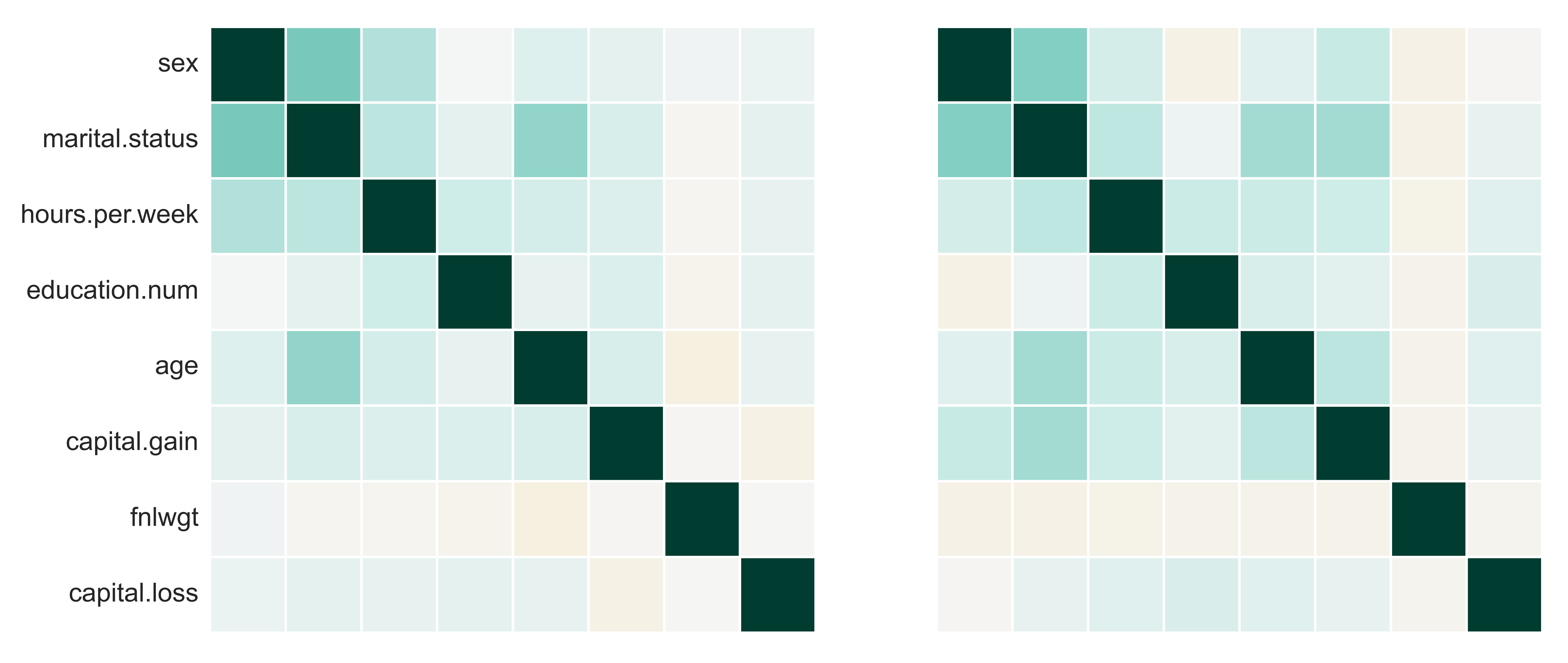}
	\caption{$\mathbf{R}^T,\mathbf{R}^S$ matrices for the Adult dataset}
	\label{}
\end{figure}

\subsection{Predictive Power comparison}

The Information Value (IV) \cite{zdravevski_weight_2011} measures the strength of the association between each variable and the target, commonly employed in Credit Scoring for feature selection. It requires categorical variables, so the continuous ones have been binned according to the deciles of their marginal distribution. We compute two IV vectors $IV^T,IV^S$ on $D_{test}$ and $D_{synth}$ respectively, and measure their similarity by means of Pearson linear correlation:
$$ \rho(IV^T,IV^S)$$

High values ensure that $D_{synth}$ maintains the same variables ranking and relative distance between values, while specific IV numbers might differ. 

\begin{figure}[htbp]
	\centering
	\includegraphics[width=0.9\textwidth]{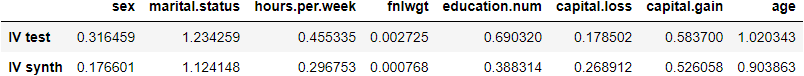}
	\caption{$IV^T,IV^S$ vectors for the Adult dataset}
\end{figure}

\section{Distributions Comparison Tests}\label{Distributions Comparison Tests}

The most powerful tool to describe a dataset is the multivariate distribution: it encodes all the variables information, such as moments, single feature marginal distribution and multivariate distribution for groups of features, pointing out correlations and interactions.\\
Having considered to estimate the $D_{test}, D_{synth}$ 	 (pdf) and subsequently compare them, we recognize this is not a viable solution. In fact, parametric estimation is not easily generalizable since it requires domain knowledge for the distribution choice, while non-parametric methods do not provide pdf formulas for the comparison. Therefore, we opted for discrepancy tests between the two pdf, without explicitly estimating them.\\
Moreover, since we deal with finite datasets, multivariate tests may fail. Thus, we detach univariate and multivariate distribution comparison. This ensures accurate results on the low-dimensional tests while it provides a theoretically solid framework to perform high-dimensional testing, even being aware they could fail due to a huge number of features.

\subsection{Univariate Distributions}
There are no proper techniques working well for both categorical and continuous features.
Convenience solutions entail to convert continuous variables into categorical, through binning methods, at the cost of loosing information (we loose granularity coercing continuous variables with many different values into a fixed number of bins). 
Conversely, it is difficult to convert categorical features to continuous without distorting the multivariate distribution. 
We keep the two comparisons separated, in order to employ more powerful tests.

\subsubsection*{Univariate Distribution - Continuous variables}

Consider the generic continuous variable $X_{(k)}$: $X_{(k)}^T$ ($X_{(k)}$ values in $D_{test}$) are drawn from the $f_{(k)}^{T}$ marginal distribution, while $X_{(k)}^S$ ($X_{(k)}$ values in $D_{synth}$) stem from $f_{(k)}^{S}$. The test goal is to assess whether the null hypothesis $\mathcal{H}_0 : f_{(k)}^{T} = f_{(k)}^{S}$ holds. 
\smallskip

Distributions are fully characterized by the collection of their moments (eg. mean, variance, skewness, kurtosis etc), hence an
intuitive approach is to search for differences in the $X_{(k)}^T,X_{(k)}^S$ moments.
 Regardless of which moment is involved, it is possible to distinguish the two distributions by choosing a proper $q$ function and evaluating its expectation. Two equal distributions shall obtain similar expected values for each valid $q$. As an example, difference in variance can be grasped using $q(x) = x^2$.

\begin{figure}[h]
\centering
\begin{subfigure}{.33\textwidth}
  \centering
  \includegraphics[width=.9\textwidth]{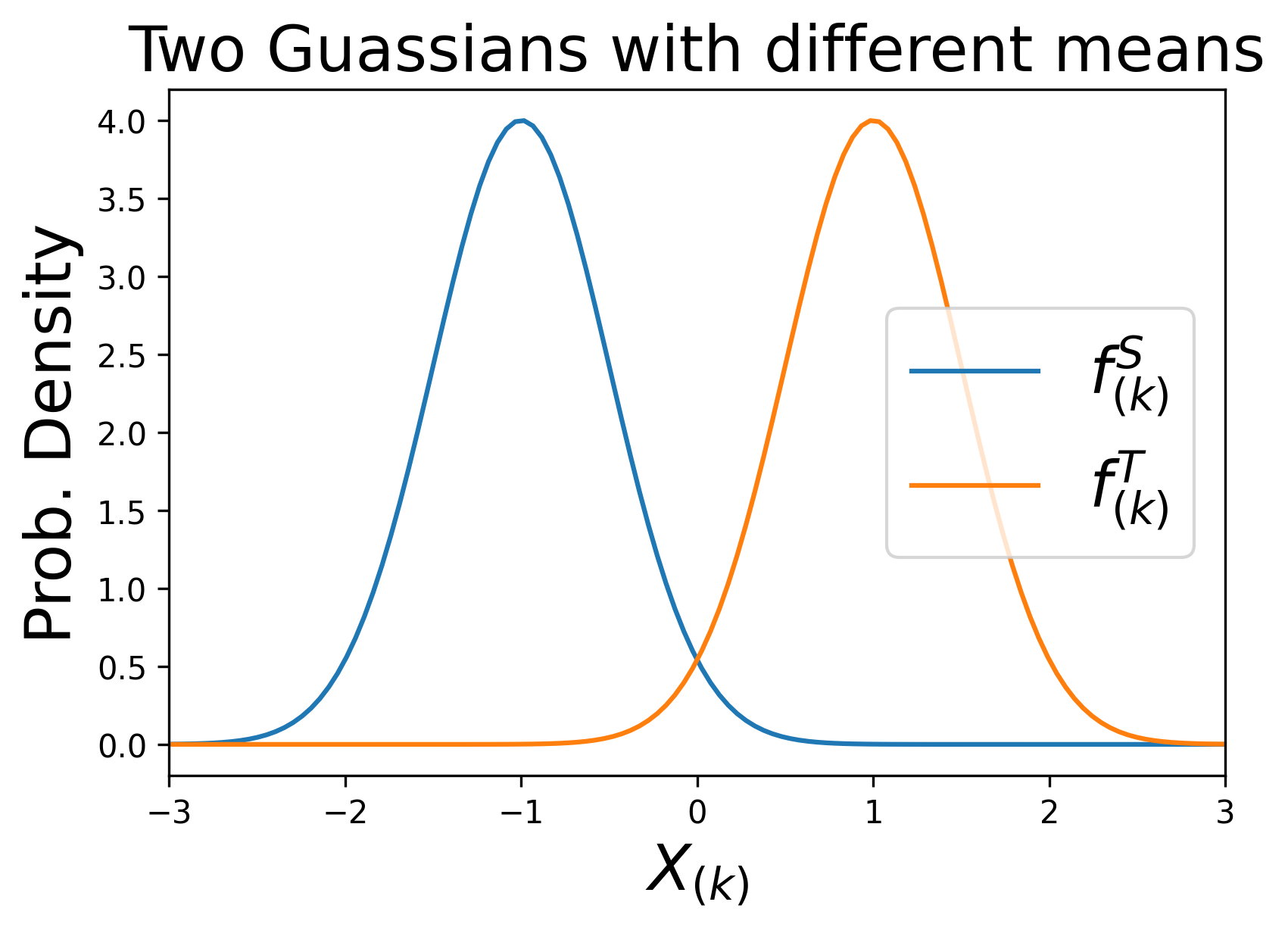}
  \caption{Difference in Mean}
  \label{fig:sub1}
\end{subfigure}%
\begin{subfigure}{.33\textwidth}
  \centering
  \includegraphics[width=.9\textwidth]{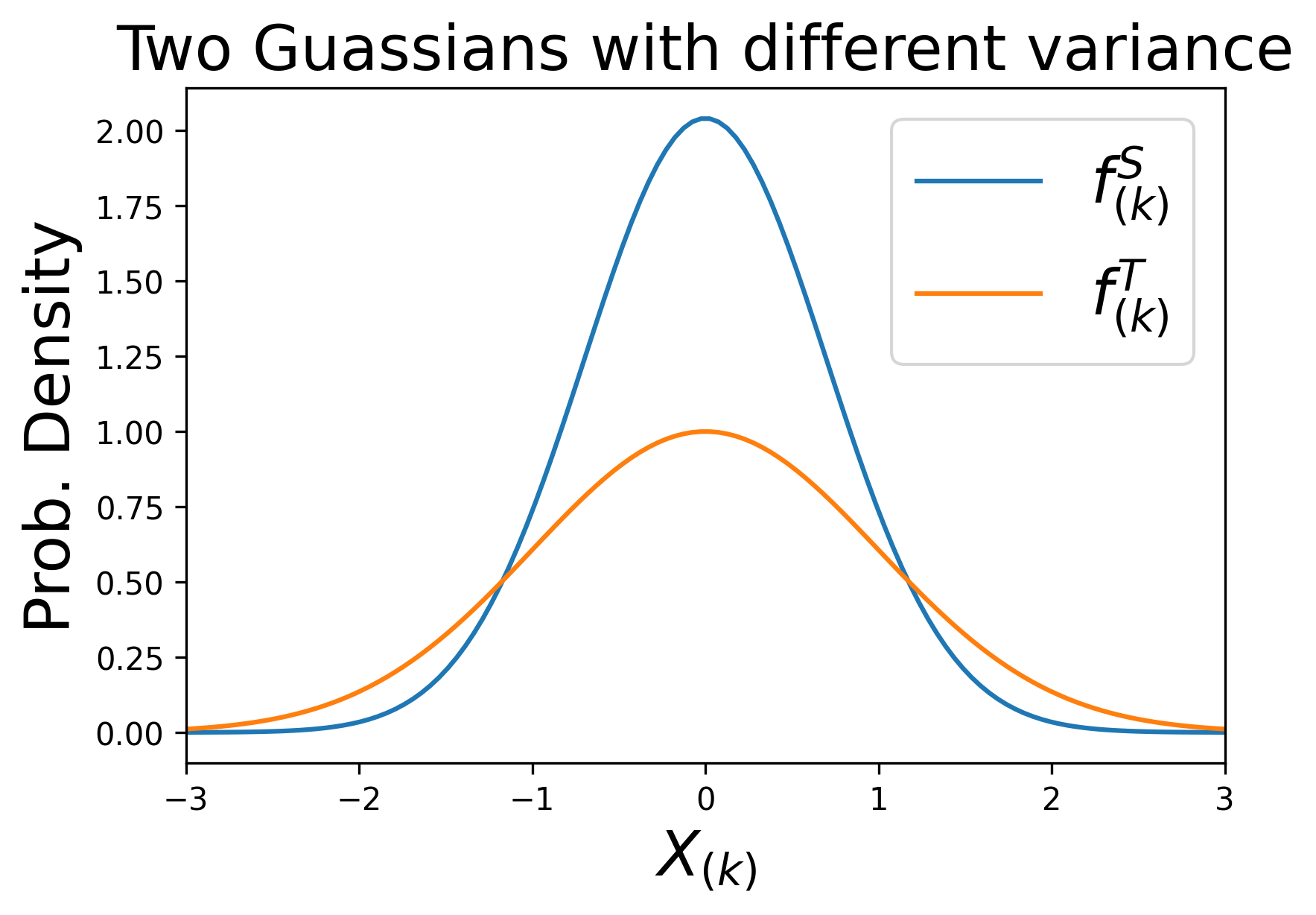}
    \caption{Difference in Variance}
  \label{fig:sub2}
\end{subfigure}
\begin{subfigure}{.33\textwidth}
  \centering
  \includegraphics[width=.9\textwidth]{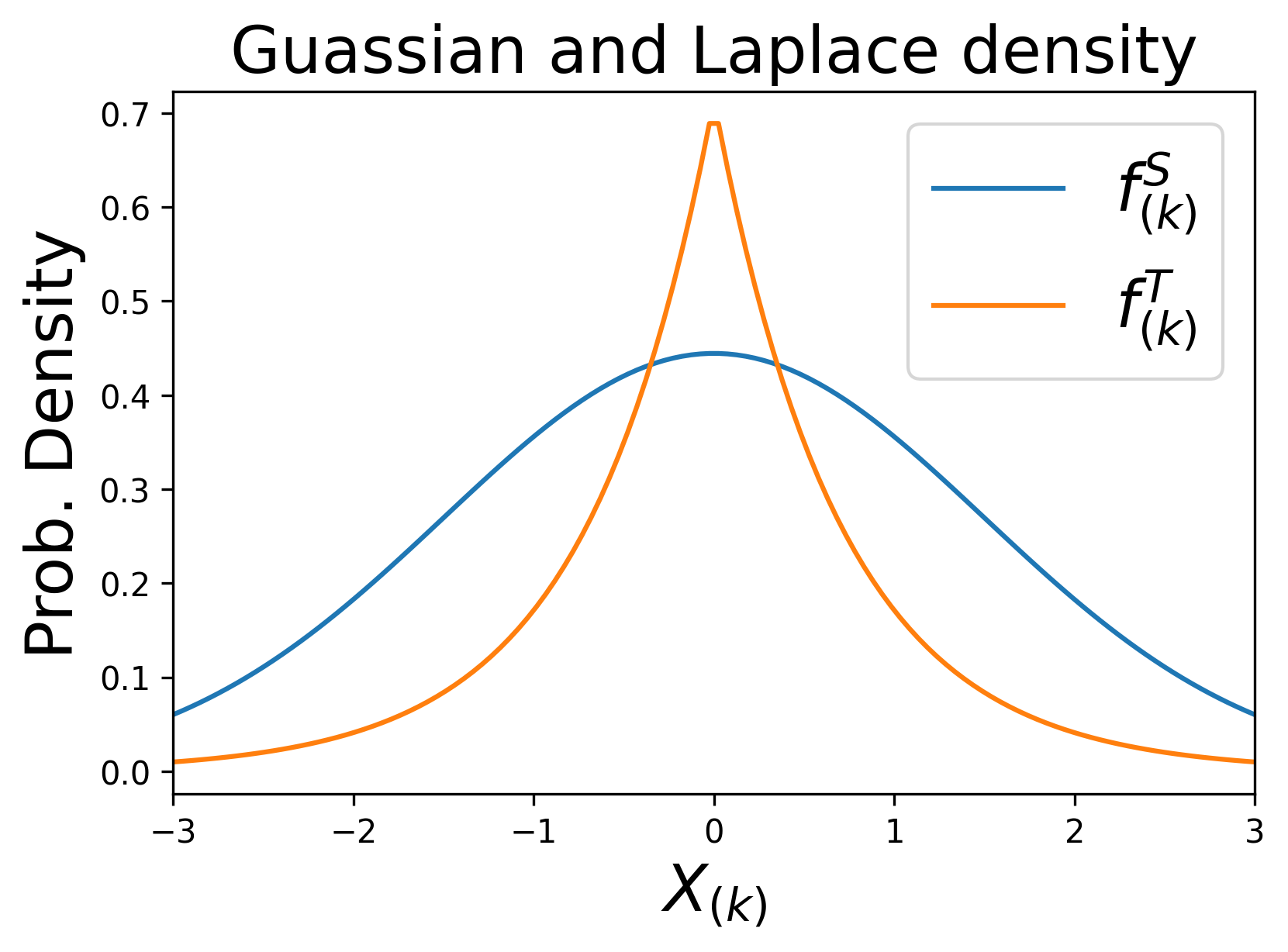}
    \caption{Difference in higher order moments}
  \label{fig:sub3}
\end{subfigure}%
\caption{Examples of different distributions}
\label{Different_distrib}
\end{figure}

Maximum Mean Discrepancy (MMD) \cite{gretton_kernel_2006} looks for the $q$ maximizing the difference between the two expectations:



\begin{equation}\label{dataset_MMD}
\operatorname{MMD}[\mathcal{Q}, X_{(k)}^{T}, X_{(k)}^{S}]:=\sup _{q \in \mathcal{Q}}\left(\frac{1}{m} \sum_{i=1}^{m} q\left(x_{(k),i}^{T}\right)-\frac{1}{n} \sum_{j=1}^{n} q\left(x_{(k),j}^{S}\right)\right)
\end{equation}

Here $m,n$ respectively stand for the $D_{test},D_{synth}$ sample size, while expectations have been replaced by the sample means. Under mild regularity conditions ($\mathcal{Q}$ contains only continuous, bounded, smooth functions), MMD is guaranteed to be different from $0$ only when $f_{(k)}^{T},f_{(k)}^{S}$ are truly different. However, the superior operator requires to screen all the valid $q \in \mathcal{Q}$, making Equation \ref{dataset_MMD} impractical to compute. An efficient MMD formulation involves kernel functions: 
\begin{equation*}
\left[\frac{1}{m^{2}} \sum_{i, j=1}^{m} k\left(x_{(k),i}^{T}, x_{(k),j}^{T}\right)-\frac{2}{m n} \sum_{i, j=1}^{m, n} k\left(x_{(k),i}^{T}, x_{(k),j}^{S}\right)+\frac{1}{n^{2}} \sum_{i, j=1}^{n} k\left(x_{(k),i}^{S}, x_{(k),j}^{S}\right)\right]^{\frac{1}{2}}	
\end{equation*}


The usual choice for $k(\cdot,\cdot)$ is the Gaussian kernel, thanks to its ability to screen infinitely large function spaces. In our implementation of the test, we take advantage of the heuristic provided in \cite{sutherland_generative_2016} to choose the proper value for $\sigma$ (the only hyper-parameter for the Gaussian kernel). \\
Eventually, we estimate the distribution of MMD under the Null Hypothesis $\mathcal{H}_0$, using a permutation test \cite{sutherland_generative_2016}. It essentially consists of randomly partitioning the data $D_{test} \cup D_{synth}$ into $D_{test}^\prime$ and $ D_{synth}^\prime$ (under $\mathcal{H}_0$ the two samples come from the same distribution) and computing the MMD. Repeating it many times, we obtain a numerical approximation of its distribution. With it, we calculate the acceptance interval (considering a 5\% type I error), through which we establish whether the variable $X_{(k)}$ passed the univariate similarity test.

\begin{figure}[h]
\centering
\begin{subfigure}{.5\textwidth}
  \centering
  \includegraphics[width=\linewidth]{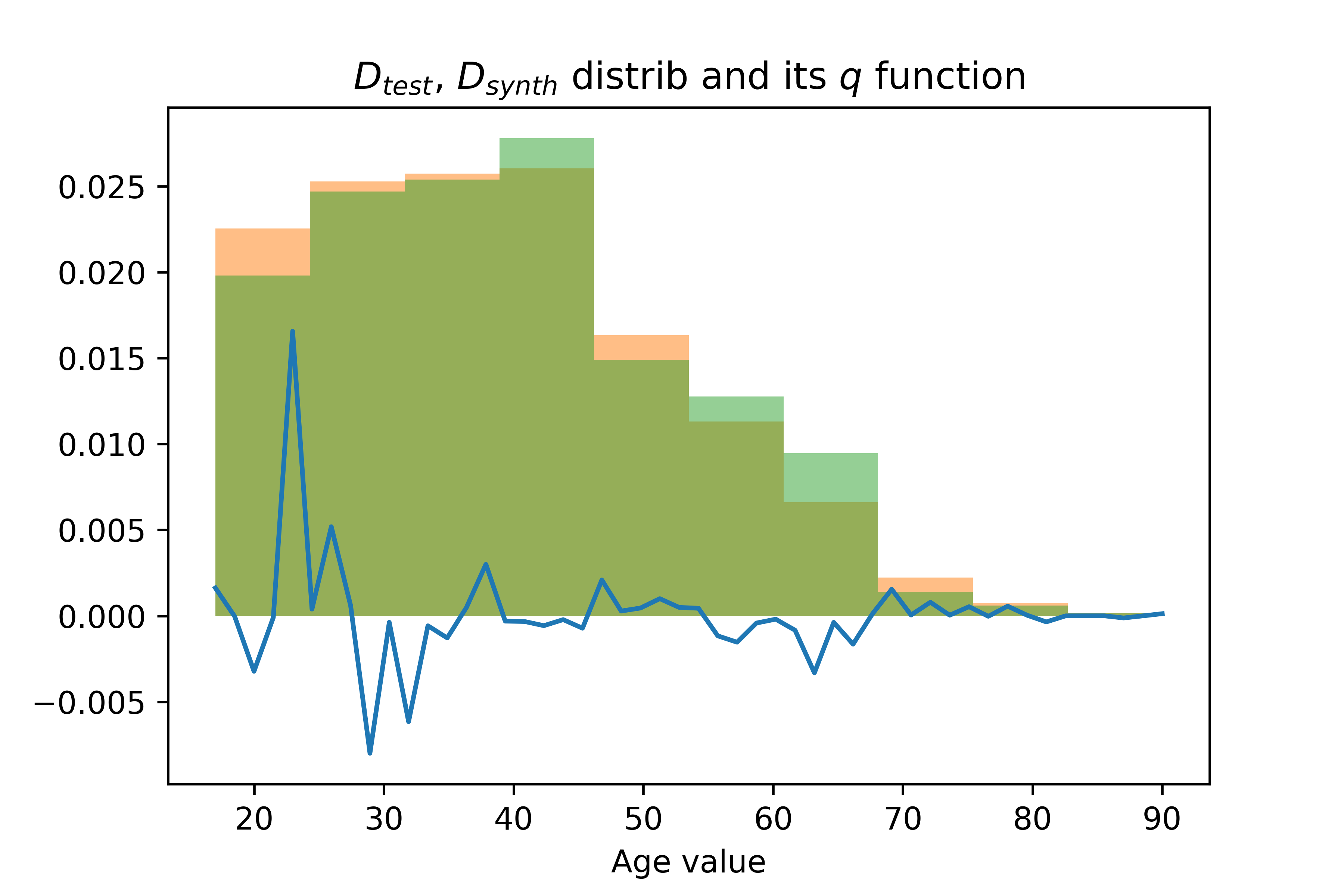}
  \caption{$D_{test},D_{synth}$ distribution histograms}
  \label{fig:sub51}
\end{subfigure}%
\begin{subfigure}{.5\textwidth}
  \centering
  \includegraphics[width=\linewidth]{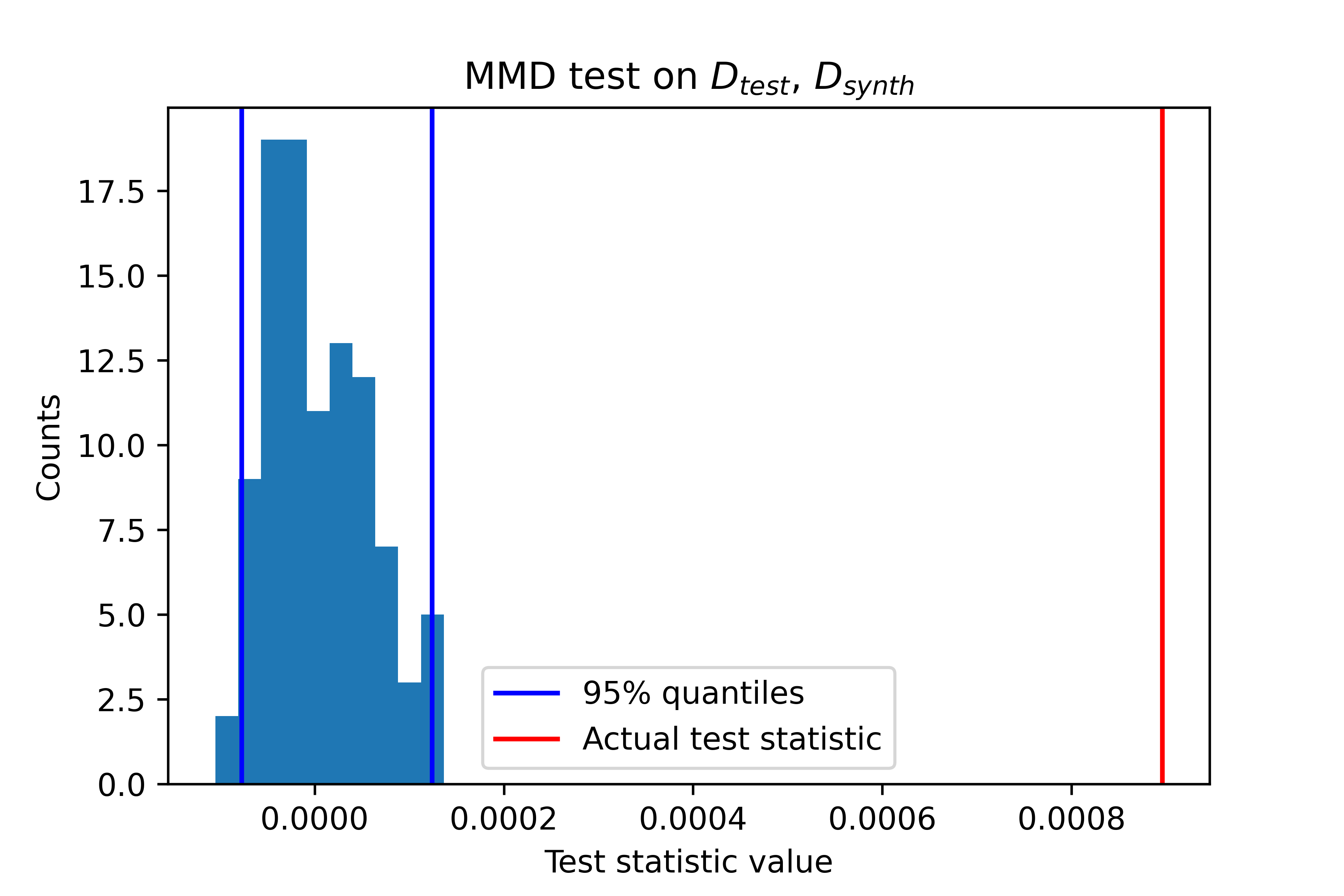}
  \caption{Confidence Interval for the MMD test}
  \label{fig:sub52}
\end{subfigure}
\caption{MMD Test on the Age variable. Here the test failed, meaning that Age distribution is different in $D_{test}$ and $D_{synth}$}
\label{our_mmd_test}
\end{figure}

While MMD is one of the most powerful tests for distribution similarity, it does not scale well for large datasets (due to kernels). In DAISYnt the test runs on $D_{test}$ and $D_{synth}$ sub-samples, whose size can be set by the user to match its computational and hardware requirements. An additional option for extremely large datasets, is to bin continuous variables using deciles and perform the faster categorical distribution testing, at the cost of a coarser result.

\subsubsection*{Univariate Distribution - Categorical variables}

The MMD based test is no use for categorical variables, therefore DAISYnt incorporates a classic Chi-Square two-sample test. The null Hypothesis $\mathcal{H}_0$ ``no significant difference in the $X_{(k)}$ class frequencies between $D_{test}$ and $D_{synth}$'' has formula: 
$$\sum_{k=1}^C \frac{\left(f_{(k),c}^{T}-f_{(k),c}^{S} \right)^{2}}{f_{(k),c}^{T}} \sim \chi^2_{(n-1) \times(C-1)}$$

where $c$ shuffles through the different $X_{(k)}$ classes, $f_{(k)}^{T},f_{(k)}^{S}$ are the class frequencies in $D_{test},D_{synth}$. Under $\mathcal{H}_0$, the test exhibits a $\chi^2$ distribution with degrees of freedom $df=(n-1) \times(C-1)$, which allows to control the type I error (DAISYnt default is 5\%). Eventually, it is important to check whether the assumptions are met, in particular the test is meaningful only if at least 80\% of the classes have an expected frequency (in this case $f_{(k)}^{T}$) greater than 5 and no class has $0$ frequency in $f_{(k)}^{S}$ \cite{fisher_testing_2011}. Following these prescriptions, DAISYnt do not test variables not complying with such criteria.
\smallskip

To obtain a unified metric, DAISYnt applies the proper test to categorical and continuous variables separately and calculates the fraction of accepted trials.

\subsection{Multivariate Distributions}

As already stated, high dimensionality and mixed data are notable challenges for two-sample distribution tests. To this end, feature selection is employed and categorical and continuous variables are considered separately. As a minor limitation, DAISYnt cannot discover any difference in distribution for groups of mixed variables.

\subsubsection*{Multivariate Distribution - Continuous Variables}

Based on empirical experiments, we deduce that MMD greatly suffers the curse of dimensionality. Hence, we restrict the similarity test to the subset of meaningful variables only, namely the relevant ones for a given prediction task. Boruta \cite{kursa_borutasystem_2010} is the method of choice, thanks to its stability, reliability and reasonable computational time. 
\smallskip

Boruta improves the Tree-based feature importance, known to be particularly unreliable in presence of correlation and interactions among variables \cite{strobl_conditional_2008}.
The method creates ``shadow features'' (variables with no relationship with the target variable) and iteratively generates Random Forest models, computing confidence intervals for the feature importance scores. Variables with significantly higher feature importance than any shadow feature are the relevant ones. Repeated trials ensure statistical stability of the procedure, while the choice of Random Forest models, which do not require any fine tuning, guarantees an acceptable computation time.
\smallskip

DAISYnt relies on Boruta to extract the set of numerical meaningful features and runs the MMD test on it. The test is much likely to fail whenever one of the important variables had already failed the univariate test. Moreover, even if all the single variables passed the univariate test, the multivariate one may fail due to differences in correlations and interactions. Substantially, the multivariate comparison is more challenging but, in case of acceptance, gives strong evidence of distribution similarity.

\subsubsection*{Multivariate Distribution - Categorical Variables}

We propose to extend the Chi-Square test to groups of categorical variables by considering a new feature, whose classes are the cartesian product of the group variables' classes, and frequency given by the contingency table. The test is carried out on the new feature. Notice that the more variables in the group, the more likely the test assumptions are not met.
\smallskip


DAISYnt considers the power set of the categorical variables.
 Starting from the 2-variables groups, it checks whether the test assumptions are fulfilled. If so, the Chi-Square test is carried out and the acceptance/rejection is recorded, otherwise the group is discarded. Bigger groups are tested only if all the sub-groups met the assumptions (otherwise the super-group would fail them either). The fraction of accepted tests is the DAISYnt similarity metric for multivariate categorical distributions. 

\subsection{Discriminator Model}

Eventually, we propose to aggregate $D_{train},D_{test}$ and $D_{synth}$ together, label each tuple as original/synthetic, and train a discriminator model to predict the label. Both $D_{train},D_{test}$ are used to ensure test robustness, for small $D_{test}$ dataset. 
Even if this test is not explicitly concerned with distributions, discriminators obtain good performance by implicitly learning distributional differences between the two datasets.
\smallskip

DAISYnt exploits a Gradient Boosting $d_{xgb}$ and a shallow Neural Network $d_{nn}$ as discriminators (more on model specifics in the Data Utility section), and evaluate their performance using Gini Index. Since we strive for synthetic data to be not distinguishable from the real ones, the discriminator test metric is rescaled as follows:
$$ 1- \frac{Gini(d_{xgb}) + Gini(d_{nn})}{2}$$

\section{Data Utility Tests}\label{Data Utility Tests}

Synthetic data have many different applications, not least to replace real data in prediction tasks. Conceptually, two datasets stemming from the same multivariate distribution shall yield the same insights, but this is not always the case when advanced models are involved. \\
To ensure synthetic data maintain data utility, DAISYnt trains prediction models respectively on $D_{train}$ and $D_{synth}$ and devises three tests with an increasing level of detail.
\smallskip

The models employed are Gradient Boosting (Xgboost \cite{chen_xgboost_2016-2} implementation) and Neural Network. Xgboost is trained with shallow decision trees of 4 splits.
The Neural Network consists of a single hidden layer of 256 neurons, with Relu activation function. Early stopping is used to avoid overfitting. Both frameworks have been trained twice on the $D_{train}$ and $D_{synth}$ datasets, obtaining four models $GB^S, GB^T, NN^S, NN^T$. DAISYnt compares the two frameworks separately, using $D_{test}$ as benchmark.

\subsection{Aggregate Prediction Comparison}

Since DAISYnt mainly focus on binary classification tasks,
it computes the AUC difference on the $D_{test}$ predictions of the four models, considered in pair.
Assuming that models trained on $D_{synth}$ cannot approximate the true underlying DGP better than the $D_{train}$ ones  (well-grounded assumption in practice), the metric is rescaled as follows:
$$1 - \frac{AUC(GB^T(D_{test})) - AUC(GB^S(D_{test}) +  AUC(NN^T(D_{test})) - AUC(NN^S(D_{test})}{2}$$

\begin{figure}[htbp]
	\centering
	\includegraphics[width=0.45\textwidth]{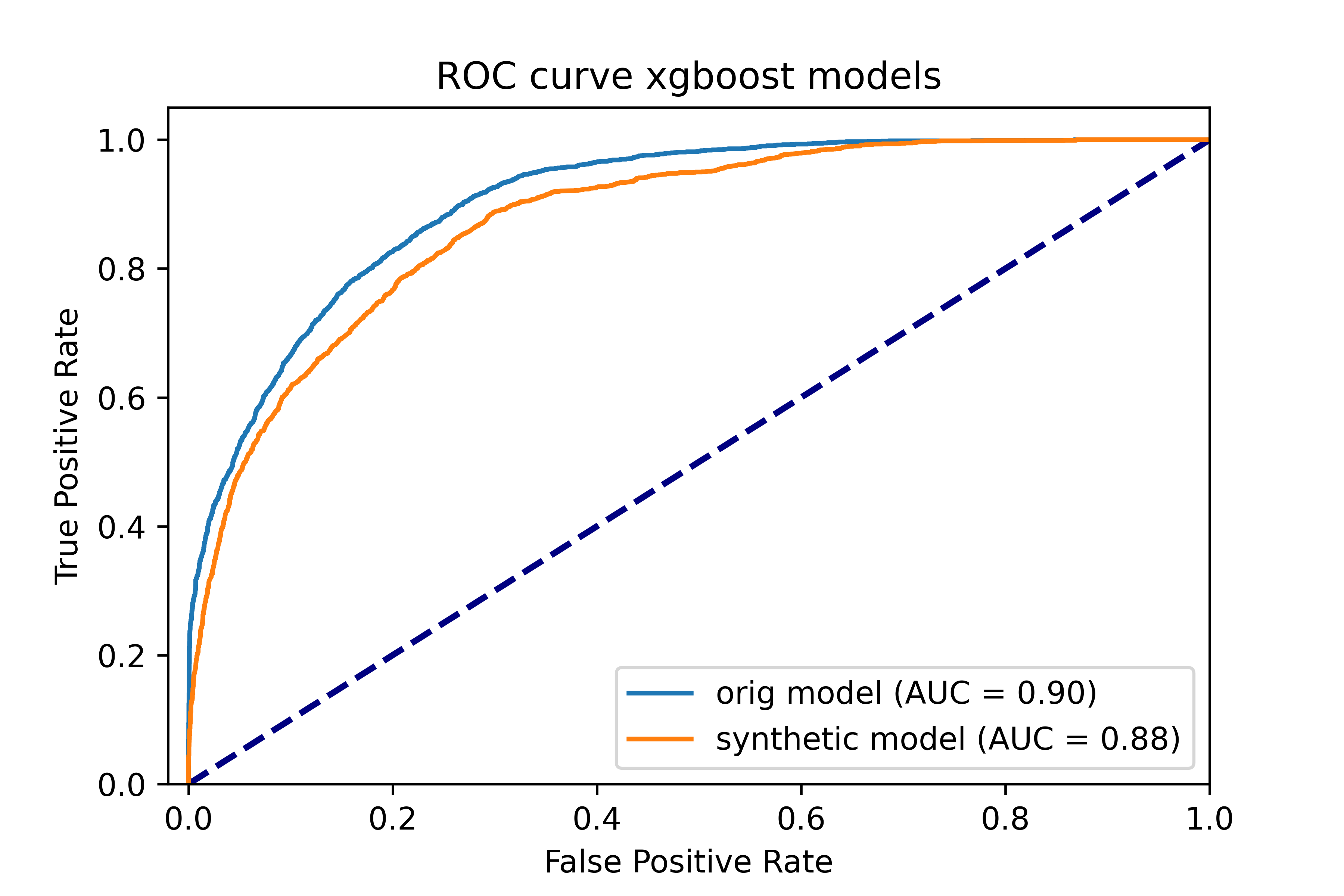}
	\caption{ROC Curve comparison of $GB^S, GB^T$ models, and relative AUC}
	\label{aggregate_prediction_comparison}
\end{figure}

\subsection{Single Prediction Comparison}

Woefully we acknowledge that AUC, due to its aggregate nature, do not guarantee the similarity of single predictions.
Conversely, obtaining the same prediction probability rankings (on the $D_{test}$ individuals) is important to produce the same insights. 
\smallskip

To this end, we define the $D_{test}$ target variable as $Y$ and the models predictions as $\hat{Y}_{GB}^S, \hat{Y}_{GB}^T, \hat{Y}_{NN}^S, \hat{Y}_{NN}^T$. 
DAISYnt compares prediction vectors using the Cosine Similarity, which takes into account solely vector direction (corresponding to the ranking concept for predictions).
Cosine similarity is evaluated on the prediction vectors, for Gradient Boosting and Neural Network separately. Results are then aggregated into a final test metric:

\begin{equation*}
	\frac{CS(\hat{Y}_{GB}^S, \hat{Y}_{GB}^T) + CS(\hat{Y}_{NN}^S, \hat{Y}_{NN}^T)}{2}
\end{equation*}


\subsection{Compare model internals}

The most refined and challenging comparison step is to check whether the model internals are the same. In a Neural Network, data flows through the network and gets transformed according to the neurons weights. We define layer activations as the values obtained by passing a data matrix through the network up to the chosen layer. The objective is to compare the activations of the $NN^T$ and $NN^S$ hidden layers, when passing $D_{test}$ as the data matrix. However, Neural Networks
 employ a huge number of parameters that frequently cause an over-representation, i.e. there are possibly infinite ways of achieving the same prediction, via different intermediate layer values. We cannot compare activations directly: they would always be different. Rather, we consider two activations to be equal when they are isotropic scaling or orthogonally invariant.

\begin{figure}[htbp]
	\centering
	\includegraphics[width=0.45\textwidth]{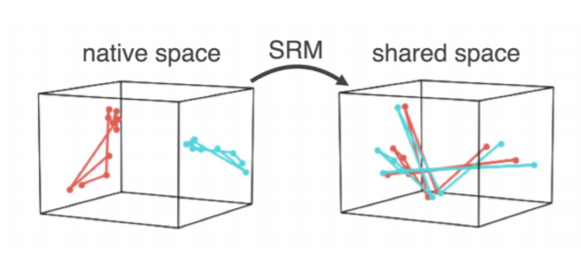}
	\caption{Toy example of transformations applied to activations matrices \cite{kornblith_similarity_2019}}
	\label{compare_model_internals}
\end{figure}

The HSIC quantity \cite{gretton_measuring_2005} captures orthogonal transformations but cannot handle the Isotropic Scaling invariance, which can be achieved by rescaling it into the Centered Kernel Alignment (CKA) \cite{kornblith_similarity_2019}. Both CKA and HSIC employ kernels to achieve the required transformations. Popular options are the linear and gaussian kernels, where the difference consists in a trade-off between flexibility and computation time. 
\smallskip

DAISYnt default consists in linear CKA between the $NN^T$ and $NN^S$ activations matrices, since the linear kernel usually achieves very similar results to the gaussian one \cite{kornblith_similarity_2019}.

\section{Privacy Tests}\label{Privacy Tests}

Since there exist no privacy regulation yet on synthetic data, we consider the taxonomy of privacy risks related to personal data, issued in the Article 29 Working Party Opinion 05/2014. Only the Linkability risk definition is slightly modified, to emphasize that building a generative model implies privacy obligations towards the $D_{train}$ individuals. The privacy dangers are grouped as: 


\begin{itemize}
    \item \textbf{Singling out risk}: The attacker single out an individual, using $D_{synth}$.

    \item \textbf{Linkability risk}: The attacker learns whether a given $D_{train}$ record has been used to train the generative model.
    \item \textbf{Inference risk}: The attacker infer the value of a sensitive attribute for new records, using the relationships contained in the $D_{synth}$.
\end{itemize}

\smallskip

Hereafter, the focus is on black-box privacy attacks (the attacker has access to synthetic data only) and how to defend against them, rather than on formal privacy frameworks such as  k-anonymity, l-diversity or differential privacy \cite{bellovin_privacy_2019}, which are difficult to evaluate post-hoc.
Membership Inference attacks usually exploit an overfitted generative model which creates $D_{synth}$ records too close or too similar to the $D_{train}$ ones. Closeness is used in \cite{hilprecht_monte_2019}, while  the similarity in \cite{song_privacy_2019}, to retrieve which records belong to $D_{train}$. A different attack can be devised using the same procedure as \cite{kassem_differential_2019}, i.e. training a model to predict a $D_{synth}$ sensitive variable and exploit it to infer confidential information on new records.\\
In the following we will present tests useful either to ensure that privacy is not at risk or to spot potential vulnerabilities.

\subsection{Singling Out Tests}

\subsubsection*{Cloned rows test}
 
The first test checks the presence of equal records in $D_{train}$, $D_{synth}$ and measures their percentage. Cloned rows can be removed from the synthetic data, at the cost of a having less data and modifying the statistical properties of the dataset.

\subsubsection*{Close rows test}

The second test looks for very similar rows. DAISYnt converts continuous variables into categorical through binning, on both $D_{train}$ and $D_{synth}$, and use the new categorical versions to calculate the Hamming distance for each original-synthetic pair of records. Pairs with Hamming distance $< 2$ are considered close. The test metric is the fraction of $D_{synth}$ records which are close to no $D_{train}$ records.



\begin{figure}[htbp]
	\centering
	\includegraphics[width=0.8\textwidth]{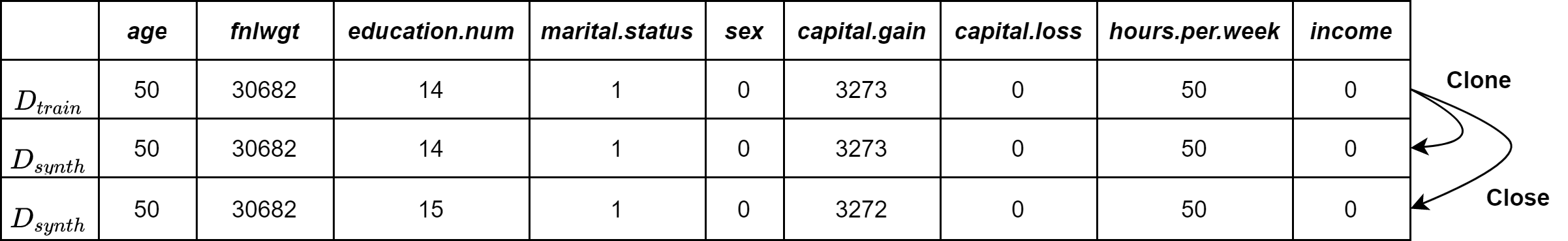}
	\caption{Cloned and close rows between $D_{train}$ and $D_{synth}$}
	\label{hamming_distance}
\end{figure}

\subsection{Linkability Tests}

An overfitted $G$ model may potentially leak information about the dataset used for training. In practice, an attacker may re-identify the $D_{train}$ records starting from $D_{synth}$, if she has access to a database containing some of the $D_{train}$ individuals.

\subsubsection*{Linkability distance test}

The assumption here is that an overfitted $G$ model shall produce $D_{synth}$ units much closer to the $D_{train}$ ones than to $D_{test}$ \cite{hilprecht_monte_2019}, i.e. choosing a fixed tiny size $\epsilon$, the $\epsilon$-neighbourhood of a generic $D_{train}$ individual $U_{\epsilon}(x \in D_{train})$ will presumably contain more  $x^\prime \in D_{synth}$ than $U_{\epsilon}(x \in D_{test})$. Since distance notions suffer the curse of dimensionality, we perform Factor Analysis of Mixed Data (FAMD) \cite{saporta_simultaneous_1990} to retrieve a restricted set of orthogonal variables. On the new coordinates system, we compute the euclidean distance matrix between $D_{train} \cup D_{test}$ and $D_{synth}$. The ``median heuristic'' \cite{hilprecht_monte_2019} supports the choice of $\epsilon$. Per each $x \in D_{train} \cup D_{test}$ we count the fraction of $D_{synth}$ units belonging to its neighbourhood, and use it as a ranking measure.
$$rank(x) = \frac{1}{n_{synth}} \sum_{i=1}^{n_{synth}} \mathbf{1} \left( x^\prime \in U_{\epsilon}(x) \right) \qquad \text{where} \;\, x^\prime \in D_{synth}$$
where $\mathbf{1}$ is the indicator function.\\
Under the initially stated assumption, $D_{train}$ data should have higher $rank$ values, i.e. the  $rank$ variable should discriminate well between $D_{train}$ and $D_{test}$ data. Its discrimination power is measured through AUC, obtaining the test metric:

\begin{equation*}
    1 - AUC_{rank}
\end{equation*}

From a different standpoint, we create a classification model exploiting the closeness of synthetic data to classify the units as $D_{train}$ or $D_{test}$.

\subsubsection*{Linkability ML test}

Information contained in $D_{synth}$ should be more valuable for re-estimating the $D_{train}$ data rather than $D_{test}$, when $G$ is an overfitted model \cite{song_privacy_2019}. To test such assumption, we build a Neural Network on the $D_{synth}$ target variable and employ the model to obtain target prediction $\hat{y}|x$ on the $x$ units of $D_{train}$ and $D_{test}$. Considering a classification task, the cross-entropy error $Err(\hat{y}|x)$ per each single prediction is computed (yet the generality of the framework allows for any other error metric to be used).\\
As before, we test the the $Err$ variable discrimination power, i.e. how well it separates $D_{train}$ from $D_{test}$ units. Considering the worst-case scenario, the threshold $\tau$ is chosen to maximize the proportion differences of $D_{train},D_{test}$ over and below $\tau$ (an attacker usually does not have enough information to choose the best $\tau$). The test metric is computed as follows:
\begin{equation*}
    1- \left( \frac{\sum_{x \in D_{train}} \mathbf{1} \{ Err(\hat{y}|x) < \tau \}}{n_{train}} - \frac{\sum_{x \in D_{test}} \mathbf{1} \{ Err(\hat{y}|x) < \tau \}}{n_{test}} \right)
\end{equation*}

\subsection{Inference Risk test}

We assume that an attacker has access to a set of public variables $X_{pub}$, but only an aggregate knowledge (average) of a sensitive variable $y_{sens}$, i.e. $\bar{y}_{sens}$. \\
We measure the $y_{sens}$ increased disclosure the attacker gains, by obtaining $D_{synth}$. The attacker may build a prediction model $f: \mathcal{X}_{pub} \rightarrow \mathcal{Y}_{sens}$ to predict the sensitive attribute on any individual, rather than predicting $\bar{y}_{sens}$.
\smallskip

Recall that $R^2$ metric calculates the prediction improvement using the $f$ model, against the basic average $\bar{y}_{sens}$ prediction. We employ a slightly modified $R^2$ version to measure the inference risk on the $D_{train}$ units:
\begin{equation*}
\frac{ \sum_{i=1}^{n_{train}} \left ( y_{i} - f(X_{pubs,i}) \right )^{2}}{ \sum_{i=1}^{n_{train}} \left ( y_{i} - \bar{y}_{sens} \right )^{2}}
\end{equation*}

Such test is valuable for the data owner to understand the dangers of sharing $D_{synth}$ and decide how to protect against them.

\section{Applications}\label{Application}

Various financial applications resort to prediction models to gain important insights. A prominent example is Credit Scoring, which consists in estimating the probability that a debtor will not repay the due amount \cite{visani_statistical_2022}.
 Several demographic, economic and financial variables concur to classify each individual as good or bad payer. 
The current application employs anonymized real-world Credit Bureau data, composed of 16 variables and $1.49\%$ of "bad payers", split into $D_{train}$ and $D_{test}$ of size $138.273$ and $59.261$ respectively. The scope of the application is to create a reliable synthetic replica to be used for data sharing with model development purposes. To this end, four generative models have been trained to produce $D_{synth}$ datasets of the same $D_{train}$ size, and DAISYnt is used to determine the best replica.
\smallskip

Different open-source \cite{patki_synthetic_2016} model implementations have been employed: i) Gaussian Copula (GC), ii) Conditional Tabular GAN (CTGAN), iii) CopulaGAN, iv) Tabular Variational Auto-Encoders (TVAE). Out-of-the-box SDV models were used with no fine-tuning of the hyper-parameters.
Concerning the test suite, DAISYnt package\footnote{\url{https://pypi.org/project/daisynt/}} 
requires only few essential parameters, namely  the $D_{train}, D_{test}, D_{synth}$ datasets, knowledge of the categorical features and the target variable. These information are used for a shallow preprocessing step, in which DAISYnt takes care of missing and special values (if specified). On categorical variables they are handled using ad-hoc classes, while on continuous variables they are treated through mean imputation and a flag indicating which records held the missing/special fields. The target variable is instead required for data utility. Additionally, we may specify the list of tests to perform, by default DAISYnt runs the entire set.
\smallskip

DAISYnt results are summarised in Table \ref{table_results}. All the tests described in the methodological sections have been performed, apart from the Inference Risk test. The latter is especially valuable to understand privacy implications of sharing synthetic data, although it requires to know which features are available to the adversary to carry out an Inference attack.


\begin{table}
\caption{DAISYnt results on the four $D_{synth}$ datasets models}\label{table_results}
\begin{center}
\begin{tabular}{c:c:c:c:c:c:c}
\toprule
\textbf{Test} &\,\textbf{Group}\, &\;\,\textbf{Detail}\;\,
&\;\textbf{GC}\; &\,\textbf{CTGAN} \, &\,\textbf{CopGAN}\, &\textbf{TVAE}\\
\bottomrule
\makecell{Correlations}&\makecell{General}&basic&0.93&0.96&0.97&0.94\\
\hdashline
\makecell{Predictive Power}&\makecell{General}&basic&0&0.97&0.92&0.38\\
\hdashline
\makecell{Uni Distrib}\makecell{(bins)\\(MMD)}&\makecell{Distrib}&\makecell{basic\\in-depth}& \makecell{0.81\\0.13}&\makecell{0.94\\0.25}&\makecell{0.88\\0.13}&\makecell{0.94\\0.25}\\
\hdashline
\makecell{Multi-Categorical \\ Distrib}&Distrib&basic&0.99&1&0.99&1\\
\hdashline
\makecell{Multi-Continuous \\ Distrib}&Distrib&in-depth&0&0&0&0\\
\hdashline
\makecell{Discriminator}&Distrib&in-depth&0.01&0.07&0.08&0.05\\
\hdashline
\makecell{Aggregate Predictions \,}&Utility&basic&0.68&0.92&0.92&0.76\\
\hdashline
\makecell{Single Predictions}&Utility&in-depth&0.01&0.46&0.46&0.09\\
\hdashline
\makecell{Model Internals}&Utility&in-depth&0.67&0.79&0.72&0.60\\
\hdashline
\makecell{Cloned Rows}&Privacy&basic&1&0.99&0.99&0.99\\
\hdashline
\makecell{Close Rows}&Privacy&basic&1&0.99&0.99&0.99\\
\hdashline
\makecell{Linkability Distance}&Privacy&basic&0.95&1&0.98&1\\
\hdashline
\makecell{Linkability ML}&Privacy&basic&1&1&1&0.99\\
\bottomrule
\end{tabular}
\end{center}
\end{table}

General purpose tests guarantee same correlation and predictive power patterns, proving that the datasets can be used for descriptive analysis. Concerning the distribution related tests, we notice satisfactory results for the univariate distributions (with binned variables) and multivariate categorical distributions. However, more in-depth tests, such as the MMD univariate, continuous multivariate distributions and the discriminator test, show quite poor values. Since we tested just the vanilla implementations, distribution discrepancies were expected. However, these results suggest the replicas cannot be employed for advanced statistical analyses, such as rebalancing good-bad payers \cite{cascarino_explainable_2022}. About data utility, aggregate predictions testify similar AUC values, but this is not enough to enable model development on the synthetic data. In fact, the more in-depth single prediction test shows that models on $D_{train}$ and $D_{synth}$ achieve substantially different insights. Eventually, high privacy metrics guarantee that privacy is preserved and the models do not overfit.\\
In conclusion, GAN models perform visibly better with respect to Gaussian Copula and TVAE, especially in terms of data utility. The generated datasets may be used for data sharing purposes, thanks to good privacy results. For model development purposes,  fine-tuning of the hyper-parameters or more powerful generative algorithms are viable solutions to achieve satisfying results. 


\section{Conclusions}\label{Conclusions}

Recent advances in the generative models field seem to solve the emerging need of high quality and private data. But the adoption of this new technology require consistent and detailed metrics to evaluate salient properties of the generated datasets. The present paper fills the gap, providing a taxonomy of the important aspects and powerful tests to assess them. The DAISYnt test suite is an easy-to-use python package, characterized by modularity and flexibility. The tests are grouped by different properties and level of detail. Based on the data generation purpose, we may choose the most appropriate tests to run. In fact, different use-cases require different quality levels of each concept. As an example, data sharing use-cases usually require high privacy levels, model development needs good data utility, while data augmentation compels strong distribution similarity. In this regard, DAISYnt is particularly versatile and business oriented.
\smallskip

Future directions entail tests refinement and extending DAISYnt to regression and multi-class classification prediction tasks,  
maintaining the applied use-cases focus and the current ease of use.
%
%
%
\bibliographystyle{splncs04}
\bibliography{Synthetic_Data}






\end{document}